\documentclass{article}
\pdfoutput=1

\usepackage[final]{neurips_2020}

\usepackage[utf8]{inputenc} 
\usepackage[T1]{fontenc}    
\usepackage{booktabs}       
\usepackage{amsfonts}       
\usepackage{nicefrac}       
\usepackage{microtype}      
\usepackage{graphicx}
\usepackage{tabularx}
\usepackage{makecell}
\title{\emph{DeepSentinel} \normalsize{ \\ \phantom{gh}} \large{ \\ An extensible corpus of labelled Sentinel-1 and -2 imagery and general purpose sensor-fusion semantic embedding model.}}

\author{%
  Lucas Kruitwagen\thanks{Contact: \texttt{lucas.kruitwagen@smithschool.ox.ac.uk}; Bio: \texttt{https://lkruitwagen.github.io/} }\\
  Smith School of Enterprise and the Environment, School of Geography and the Environment\\
  University of Oxford\\
  Oxford, UK \\
}

\begin{document}

\maketitle

\begin{abstract}
  Earth observation offers new insight into anthropogenic changes to nature, and how these changes are effecting (and are effected by) the built environment and the real economy. 
  With the global availability of medium-resolution (10-30m) synthetic aperature radar (SAR) Sentinel-1 and multispectral Sentinel-2 imagery, machine learning can be employed to offer these insights at scale, unbiased to the reporting of companies and countries. 
  In this paper, I introduce \textit{DeepSentinel}, a data pipeline and experimentation framework for producing general-purpose semantic embeddings of paired Sentinel-1 and Sentinel-2 imagery. 
  I document the development of an extensible corpus of labelled and unlabelled imagery for the purposes of sensor fusion research. 
  With this new dataset I develop a set of experiments applying popular self-supervision methods and encoder architectures to a land cover classification problem.
  Tile2vec spatial encoding with a self-attention enabled ResNet model outperforms deeper ResNet variants as well as pretraining with variational autoencoding and contrastive loss.
  All supporting and derived data and code are made publicly available.
\end{abstract}

\section{Earth observation for climate change mitigation}
Satellite-based earth observation plays a central role in measuring climate change impacts and risks \cite{caldecott2018}.
Medium-resolution satellites (10-30m spatial resolution), despite being initially designed for environmental monitoring, are being increasingly used in applications focusing on the interface between the environment and the real economy for the purposes of financial risk measurement: estimating carbon dioxide emissions\cite{cti18}, methane emissions\cite{varon2020}, and localising large fixed-capital assets\cite{kruitwagen19}.
For the purposes of assessing financial risk due to climate change, these satellites provide the globally exhaustive and unbiased view required by financial decision makers.
Deploying analysis at this global scale can only be accomplished with the use of machine learning.

These climate change risk applications are impaired by two perennial challenges with satellite-based earth observation: the presence of atmospheric interference, and a shortage of training labels.
Atmospheric interference is not equally distributed around our planet. 
Excessive cloud cover makes surface retrievals using multispectral imagery very challenging in certain geographies, negating its otherwise `exhaustive` coverage.
Many cloudy geographies are in the global south, precisely where populations are the most vulnerable to climate change and where conventional reported data is the most sparse.
These same geographies are where financial institutions in the global north concentrate their risk exposure to generate outsize returns, and are where civil society groups must be most vigilant to identify and respond to neocolonial practises by these same institutions.

The shortage of training labels for machine learning with earth observation data is well documented.
Land use and land cover data is available with moderate spatial and temporal resolution in Europe (e.g. the Copernicus Corine Land Cover\cite{corine:20}) and the US (e.g. the USDA Cropland Data Layer \cite{usda:20}), however data for the rest of the planet are sparse (e.g. OpenStreetMaps \cite{osm}).
These datasets include only broad categories of land use and land cover, and are not fit for the purpose of localising specific categories of industrial infrastructure.
Financial institutions just beginning to reckon with geospatial data, leaving a large gap in existence and availability of spatially-localised asset-level data.\cite{spglobal20}

\begin{figure}
\centering
\includegraphics[angle=90,width=0.7\textwidth]{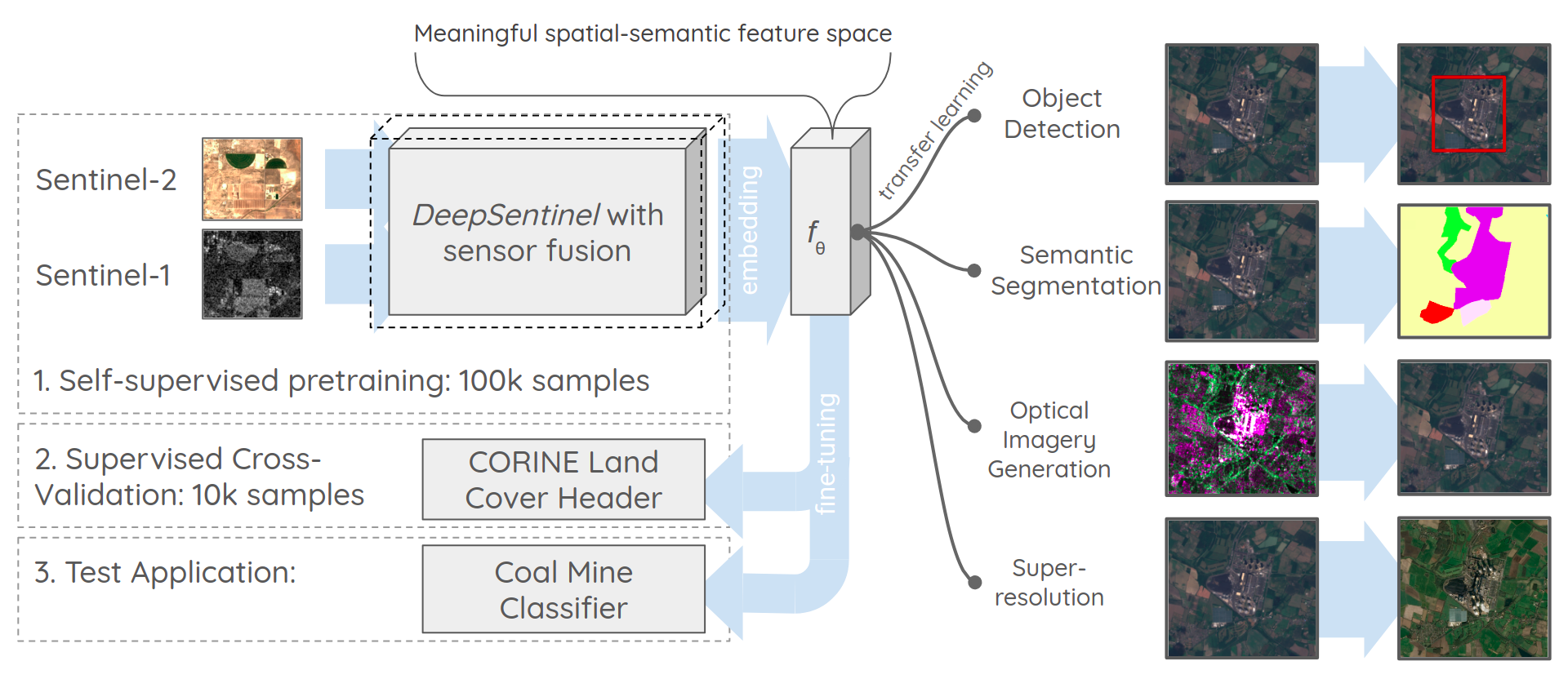} 
\caption{
\emph{DeepSentinel} summary, showing pretraining and fine-tuning curriculum and the variety of use case applications.
}
\label{fig:0-deepsentineldesign}
\end{figure}

I develop a general-purpose sensor fusion semantic embedding model to overcome these dual challenges of atmospheric interference and label availability. 
\emph{DeepSentinel} uses sensor fusion of SAR Sentinel-1 data and multispectral Sentinel-2 data to provide latent space embeddings of surface conditions even in excessively cloudy conditions.
Self-supervised pretraining followed by fine tuning on land-use and land-cover data creates a general purpose embedding model suitable for a wide range of downstream applications using transfer learning, see Figure \ref{fig:0-deepsentineldesign}.
This paper provides a background on the current state-of-the-art in sensor fusion, documents a data corpus for training purposes, and describes experimentation and initial results towards \emph{DeepSentinel}, a general-purpose embedding model.

\section{Sensor fusion}
SAR imagery carries complementary information to multispectral imagery and so is useful beyond its penetration of atmospheric conditions. 
Sentinel-1 SAR C-band backscatter is sensitive to moisture and surface types, and has applications in both natural environments (e.g. with classification of forests\cite{ruetschi2018} and crop types\cite{nelson2014,setiyono2017,ndikumana2018}), and the built human environment (e.g. road classification \cite{zhang2019}).
These properties are complimentary to the multispectral data provided by Sentinel-2, which has sensors designed to detect aerosols, water vapour, chlorophyll, and visual spectra.

Sensor fusion describes the complementary combination of data from multiple sources to improve inference quality beyond what would be possible from either sensor individually.
Fusion of Sentinel-1 and Sentinel-2 imagery has been studied for cloud removal\cite{eckardt2013,grohnfeldt2018,bermudez2018,meraner2020}, synthetic imagery generation \cite{he2018,bermudez2019,fuentes2019}, and land cover classification \cite{van2018,ferrant2017,torbick2018}.

With the exception of \citep{meraner2020}, most of these studies obtain a limited corpus for a specific area of interest.
With \emph{DeepSentinel}, we propose a general-purpose encoder than can be used for any land-surface area-of-interest on the planet, and fine-tuned for any of the applications above.
We prepare an unlabelled training dataset of similar size to \citep{meraner2020}, see Table \ref{tab:datasets}.
Our dataset differs in that it is provided corrected for atmosphere conditions, and so shows the best-available surface reflectance retrievals.
It has not been generated with any bias against cloud cover and has been uniformly sampled over the planet's land surface area, so it represents raw, albeit preprocessed, data from the natural distributions of the two sensing platforms.
The data have been generated twice, sampling from both Google Earth Engine and DescartesLabs imagery access pipelines, making our results are universally accessible as possible.
Our goal is to unlock increasingly niche applications where limited training data is available, allowing a proliferation of earth observation use cases with impact analogous to the release of pre-trained conventional imagery convolutional neural networks of the mid-2010s (e.g. ResNet, VGG-16/19, etc).

DeepSentinel includes a dataset with labels obtained from two sources: land cover labels from Copernicus CORINE Land Cover \cite{corine:20}, and polygon, line, and point data from OpenStreetMaps (OSM). \cite{osm}
Copernicus CORINE Land Cover data was rasterised from geodatabases with less than 100m spatial resolution.
This dataset was geographically restricted to the EU27+GB, where the land cover data was available and where OSM data is of significantly higher quality.
While the maximum resolution of the geodatabases is coarser than the spatial resolution of imagery, rasterising directly from the geodatabases rather than use derived 100m raster products means the actual resolution of the land cover samples is considerably higher (although not consistent across the EU27+GB countries).
In this sense these are `weak' labels which do not confidently match the true land cover at each location.

\begin{figure}
\includegraphics[width=1.0\textwidth]{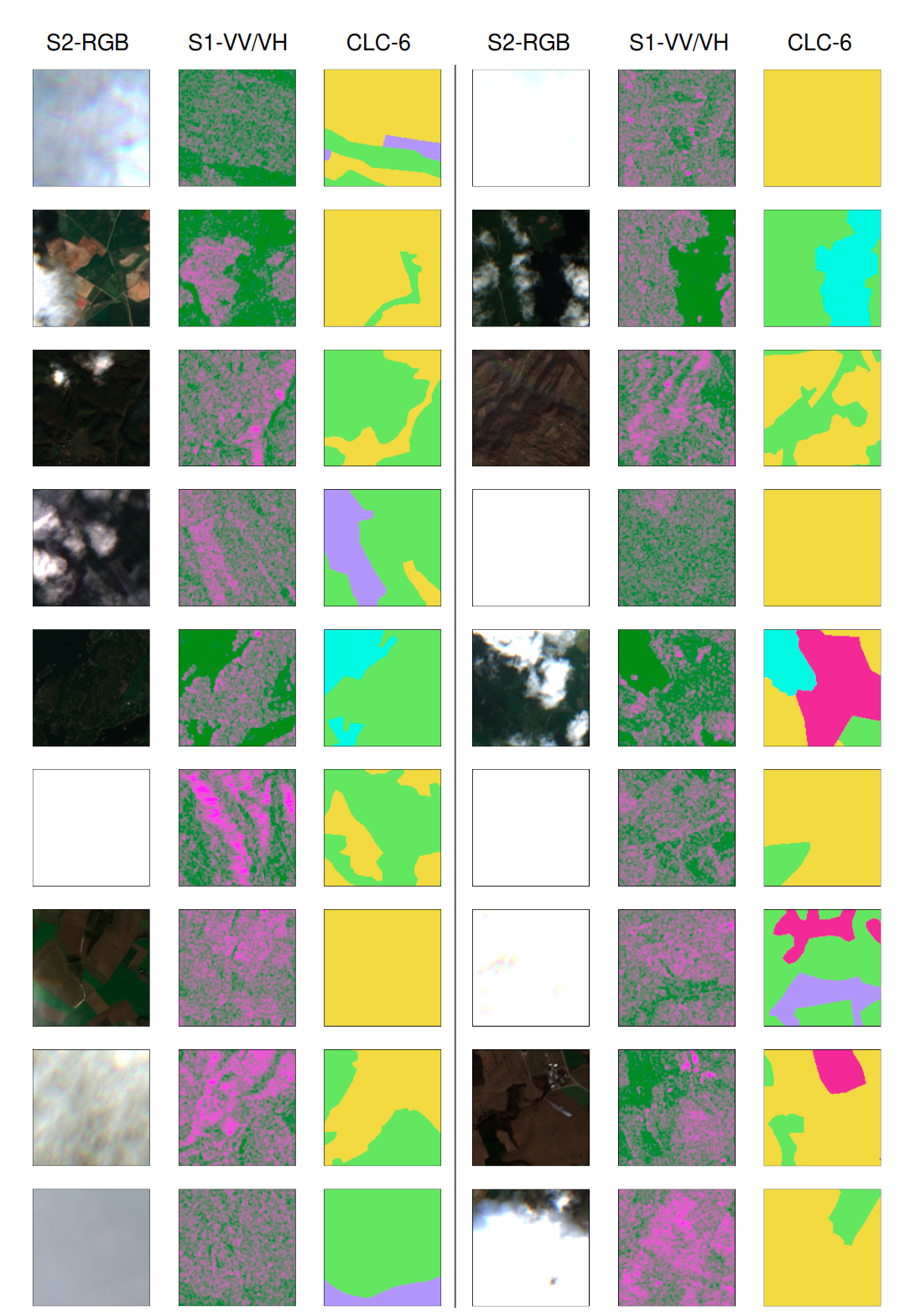} 
\caption{
Samples from the training corpus. S2: Sentinel-2; S1: Sentinel-1; CLC-6: 6-class Copernicus CORINE land cover classification. Blank Sentinel-2 RGB images area not missing data - they are cloudy images where the brightness saturates all pixel values. Sentinel-1  
}
\label{fig:1-examples}
\end{figure}

A sample of the labelled data is shown in Figure \ref{fig:1-examples}.

\begin{table}
\begingroup
\centering
\caption{Sensor Fusion Datasets}
\small
\begin{tabular}{p{2.7cm} p{4cm} p{2cm} p{1.3cm} p{1.6cm}} 
    \toprule 
      Dataset (Study) & Components* & Pixels & Resolution & Geography \\
     \midrule
     \makecell[l]{\emph{DeepSentinel} labelled \\ (ours)} & S1 (VV+VH) + S2 (L2A - all bands) + CCLC + OSM & 655 Mpx & 10m & EU27+GB \\
      \makecell[l]{SEN12MS-CR \\ (Meraner et al. 2020)} & S1 (VV+VH) + S2 (L1C - all bands) & 10,323 Mpx & 10m & 169 Global ROIs \\
     \makecell[l]{\emph{DeepSentinel} \\ unlabelled (ours)} & S1 (VV+VH) + S2 (L2A - all bands) & 6,554 Mpx * 2 & 10m & Global Uniform \\
     \bottomrule
\label{tab:datasets}
\end{tabular}
{\newline\raggedright * S1: Sentinel-1; S2: Sentinel-2; CCLC: Copernicus CORINE Land Cover; OSM: OpenStreetMaps; L1C: Level 1-C; L2A: Level 2-A}\\
\endgroup
\end{table}


\section{Data preparation}
Two novel datasets are prepared for the development of \emph{DeepSentinel}.
The first dataset is prepared without labels for the purpose of self-supervised pretraining.
Random sample patches are obtained from the planet's land surface area.
The square patches are sampled at a pixel resolution of 10m with 256-pixel side length.
Patches are obtained for both Sentinel-1 and -2, only where the image acquisition dates are within 3 days of each other.
For Sentinel-2, all 12 multispectral bands are sampled.
For Sentinel-1, VV and VH polarisation bands of interferometric wide swath (IW) retrievals are sampled, the retrieval mode used over land.
For maximum reproduceability and impact, samples for all patches are obtained from both Google Earth Engine\cite{gee17} and Descartes Labs computation platform\cite{descarteslabs17}.
Sentinel-2 acquisitions from Google Earth Engine are provided at `Level 2A' surface reflectance level, from Descartes Labs they are obtained at `surface-level' using Descartes Labs proprietary atmosphere correction algorithm. 
Sentinel-1 acquisitions from Google Earth Engine have been thermal noise corrected, radiometrically calibrated, and terrain corrected.
Sentinel-1 acquisitions from Descartes Labs have been similarly terrain corrected.

The second dataset is prepared with labels for the purposes of fine-tuning and cross-validation.
Land use and land cover labels are obtained from the 2018 Copernicus CORINE Land Cover inventory, rasterised at 10m, as well as OSM Point, Line, and Polygon datasets.
For the second dataset, samples are only obtained for European Union (plus the United Kingdom) where Copernicus CORINE Land Cover data are available and OSM data is of substantially higher quality.
The two datasets are made publicly available via Google Cloud Storage and Microsoft Azure Storage.
Four datasets have been prepared: a demonstration 1,000 sample dataset with CORINE and OSM land use and land cover labels, a 10,000 sample dataset without labels, a 10,000 dataset with CORINE and OSM land use and land cover labels, and a 100,000 sample dataset without labels.\footnote{Datasets are accessible at \texttt{https://console.cloud.google.com/storage/browser/deepsentinel}} The code for sampling the earth observation, land cover, and OSM data is available via Github. \footnote{\emph{DeepSentinel} code is available at \texttt{https://github.com/Lkruitwagen/deepsentinel.git} and the OSM server at \texttt{https://github.com/Lkruitwagen/deepsentinel-osm.git}}
With 100,000 unlabelled samples, our dataset is similar in size to the largest dataset available for sensor fusion research.\cite{meraner2020}

\section{Experimentation}
The potential of DeepSentinel is demonstrated with a set of experiments producing semantically-meaningful features embeddings from Sentinel-1 and -2 imagery.
The ultimate goal of DeepSentinel is to be able to take advantage of the abundant unlabelled imagery and limited or weak imagery labels to produce a general-purpose embedding machine prior to fine-tuning on specific end-use applications.
Such a system features two primary design decisions: the choice of self-supervision training curriculum, and a choice of embedding architecture.
This paper presents experimentation and results for three self-supervision algorithms and three embedding architectures to see which produces the best image representations for land cover classification and coal mine classification.

The three self-supervision approaches used in this paper are variational autoencoding \cite{kingma2013auto}, tile2vec\cite{jean2019} spatial encoding, and contrastive learning \cite{oord2018representation,tian2019contrastive}.
These three methods follow different intuitions in the design of self-supervision criteria. 
Variational autoencoding combines the information compression of a standard autoencoder with generative sampling of a latent space to ensure continuous and semantically-meaningful embeddings.
Both tile2vec and contrastive learning seek to minimise euclidean distances of latent space representations of `alike' samples, while maximising the distance between the representations of `distant' samples.
With tile2vec `alike-ness' is dictated by geospatial proximity.
With contrastive learning, `alike-ness' is determined by the generation of different views of randomly-augmented samples. 
Experimenting with all three of these methods will show which is most efficient at encoding high-level feature information in fused Sentinel-1 and -2 imagery.

A variational autoencoder (VAE) is prepared which encodes samples from the unlabelled dataset and generates a reconstructed image.
The autoencoder is trained using reconstruction loss -- the mean-squared error between the generated 14-channel genearted image and the input image.
Several variations are made to the architecture which draws on Ha \& Schmidhuber's \cite{ha2018world} implementation.
Rather than flattening the entire embedding space, dimension reduction is achieved by first using 1x1 convolutional kernels.
This reduces embedding space dimensions while maintaining spatial relationships.
Deconvolution of the generated latent vector uses transposed convolution followed by two standard convolution-batch-normalisation-rectified-linear-unit blocks.

Tile2vec\cite{jean2019} (T2V) is a spatial analog to word2vec\cite{mikolov2013efficient} and functions on the (Tobler-ian)\footnote{Tobler's First Law of Geography \cite{tobler1970computer}: ``everything is related to everything else, but near things are more related than distant things.''} assumption that objects that are nearby are more alike than objects that are distant.
Within the unlabelled dataset, for each point, the geodesic distance to the nearest neighbours that are within 1$^o$ is obtained.
The average distance between points (`anchors') and their closest neighbour is 18.2km.
Training uses triplet loss, where for each anchor point, the euclidean distance between the latent space representation of the anchor and neighbour points is minimised, and distance between the anchor and distant point representations is maximised. 
Neighbour points are chosen from the neighbours within 1$^o$, with probability equal to the softmax of their distances to the anchor point.
Distant points are chosen from the remaining non-neighbour population.
T2V uses the embedding model directly and has no other learned parameters.

The final self-supervision approach uses contrastive learning.
In contrastive learning, the same underlying latent space representation may have multiple possible views in the data space.
Adapting Contrastive Sensor Fusion (CSF) from \cite{swope2020representation}, paired image views are prepared where each view has been augmented with band dropout, random cropping, and contrast, brightness, saturation, and hue jitter.
A loss function is defined as the euclidean distance between the embedded representations of these augmented views.
To avoid a null solution, the distance between other randomly selected samples should also be maximised, and the ensemble is trained using triplet loss as with tile2vec.
As in \cite{swope2020representation}, a learning curriculum si employed whereby augmentation intensifies over a period of 10 epochs following a `warm-up` phase of 5 epochs.

All pretraining methods are trained on the 100,000 sample unlabelled dataset of matched Sentinel-1 and Sentinel-2 imagery. 
The imagery bands are concatenated at model input and are normalised using the band-wise means and standard deviations.
Pretraining for all methods is run for 20 epochs with a maximum batch size of 1024 on a cluster of eight NVidia k80 GPUs. 
Adam optimisers are used for training, with an initial learning rate of 5$e-4$.

The challenge with self-supervision methods is the development of an evaluation criteria suitable for a target use case.
This is why \emph{DeepSentinel} is made available with a second dataset which includes land cover labels for the EU27+GB.
Pretrained encoders are finetuned using the labelled data in a semantic segmentation land cover classification task.
Pixels are classified into the categories shown in Table \ref{tab:LCClasses}.
Cross-validation performance on the fine-tuning task is also used to optimise the self-supervised pretraining method and to develop insight into the representations the encoders are learning.

Three well-studied variants of the ResNet architecture\cite{he:15} are used to develop insight into favourable encoder properties.
ResNet18 and ResNet34 are topologically similar and differ only in their depth and number of parameters.
The difference in performance between the two models gives insight into the parameter spaces required to encode deep semantic information.
The third embedding architecture is a ResNet variant incorporating context-sensitive attention.\cite{vaswani2017attention}
Attention used in this way transports broader contextual information into the convolutional perceptive field.
The three embedding architectures are shown in Tables \ref{tab:RN18}, \ref{tab:RN34}, \ref{tab:RN18Attn}, with reference to Tables \ref{tab:downblock}, \ref{tab:upblock}, \ref{tab:attnblock}, and \ref{tab:DeconvHeader}.


\begin{table}
\begingroup
\centering
\caption{Convolutional Block Architecture (DownBlock) \protect\cite{he:15}}
\small
\begin{tabular}{p{2.5cm} p{2.5cm} p{4cm} r} 
    \toprule 
      Layer & Input & Operation & Output Shape \\
     \midrule
     Conv1 & INPUT [B,C,X,Y] & Convolution2D & [B,C*2,X/2,Y/2] \\
     BatchNorm1 & Conv1 & Batch Normalisation & [B,C*2,X/2,Y/2]  \\
     ReLu1 & BatchNorm1 & Rectified Linear Activation & [B,C*2,X/2,Y/2]  \\
     Conv2 & ReLu1 & Convolution2D & [B,C*2,X/2,Y/2]  \\
     BatchNorm2 & Conv2 & Batch Normalisation & [B,C*2,X/2,Y/2] \\
     ReLu2 & BatchNorm2 & Rectified Linear Activation & [B,C*2,X/2,Y/2] \\
     Out & [INPUT,ReLu2] & Sum & [B,C*2,X/2,Y/2] \\
     \bottomrule
\label{tab:downblock}
\end{tabular}
\endgroup
\end{table}

\begin{table}
\begingroup
\centering
\caption{Deconvolutional Block Architecture (UpBlock) \protect\cite{he:15}}
\small
\begin{tabular}{p{2.5cm} p{2.5cm} p{4cm} r} 
    \toprule 
      Layer & Input & Operation & Output Shape \\
     \midrule
     DeConv1 & INPUT [B,C,X,Y] & Transpose Convolution2D & [B,C/2,X*2,Y*2] \\
     Conv1 & DeConv1 & Convolution2D & [B,C/2,X*2,Y*2] \\
     BatchNorm1 & Conv1 & Batch Normalisation & [B,C/2,X*2,Y*2] \\
     ReLu1 & BatchNorm1 & Rectified Linear Activation & [B,C/2,X*2,Y*2] \\
     Conv2 & ReLu1 & Convolution2D & [B,C/2,X*2,Y*2] \\
     BatchNorm2 & Conv2 & Batch Normalisation & [B,C/2,X*2,Y*2] \\
     ReLu2 & BatchNorm2 & Rectified Linear Activation & [B,C/2,X*2,Y*2] \\
     \bottomrule
\label{tab:upblock}
\end{tabular}
\endgroup
\end{table}

\begin{table}
\begingroup
\centering
\caption{Attention Block Architecture (AttnBlock) \protect\cite{zhang2019self}}
\small
\begin{tabular}{p{2.5cm} p{3.5cm} p{4cm} r} 
    \toprule 
      Layer & Input & Operation & Output Shape \\
     \midrule
     QueryConv & INPUT [B,C,X,Y] & Convolution2D$_{1x1}$ & [B,C/8,X,Y] \\
     KeyConv & INPUT [B,C,X,Y] & Convolution2D$_{1x1}$  & [B,C/8,X,Y] \\
     ValConv & INPUT [B,C,X,Y] & Convolution2D$_{1x1}$  & [B,C,X,Y] \\
     Energy & [QueryConv$^T$, KeyConv] & Matrix Multiplication & [B,C/8,X,Y] \\
     AttentionMap & Energy & Softmax & [B,C,X,Y] \\
     PreOut & [ValConv, AttentionMap] & Matrix Multiplication & [B,C,X,Y] \\
     Out & [PreOut, INPUT] & Sum & [B,C,X,Y] \\
     \bottomrule
\label{tab:attnblock}
\end{tabular}
\endgroup
\end{table}

\begin{table}
\begingroup
\centering
\caption{ResNet18 \protect\cite{he:15}}
\small
\begin{tabular}{p{1.8cm} p{3.2cm} p{3.5cm} c r} 
    \toprule 
     Layer & Input & Operation & Output Shape & Parameters\\
     \midrule
     Conv1 & INPUT [B, 14, 128, 128] & Convolution2D & [B, 64, 64, 64] & 43,904 \\
     BatchNorm1 & Conv1 & Batch Normalisation & [B, 64, 64, 64] & 128 \\
     ReLu1 & BatchNorm1 & Rectified Linear Activation & [B, 64, 64, 64] & 0 \\
     Layer1 & ReLu1 & DownBlockx2 (Table \ref{tab:downblock}) & [B, 64, 64, 64] & 147,960 \\
     Layer2 & Layer1 & DownBlockx2 (Table \ref{tab:downblock}) & [B, 128, 32, 32] & 525,568 \\
     Layer3 & Layer2 & DownBlockx2 (Table \ref{tab:downblock}) & [B, 256, 16, 16] & 2,099,712 \\
     Layer4 & Layer3 & DownBlockx2 (Table \ref{tab:downblock}) & [B, 512, 8, 8] & 8,393,728 \\
     Out & Layer4 & Softmax & [B, 512, 8, 8] & 0\\
     \midrule
     \multicolumn{4}{r}{\textbf{Trainable Parameters:}} & \textbf{11,211,008} \\
     \bottomrule
\label{tab:RN18}
\end{tabular}
\endgroup
\end{table}

\begin{table}
\begingroup
\centering
\caption{ResNet34 \protect\cite{he:15}}
\small
\begin{tabular}{p{1.8cm} p{3.2cm} p{3.5cm} c r} 
    \toprule 
     Layer & Input & Operation & Output Shape & Parameters\\
     \midrule
     Conv1 & INPUT [B, 14, 128, 128] & Convolution2D & [B, 64, 64, 64] & 43,904 \\
     BatchNorm1 & Conv1 & Batch Normalisation & [B, 64, 64, 64] & 128 \\
     ReLu1 & BatchNorm1 & Rectified Linear Activation & [B, 64, 64, 64] & 0 \\
     Layer1 & ReLu1 & DownBlockx3 (Table \ref{tab:downblock}) & [B, 64, 64, 64] & 221,952 \\
     Layer2 & Layer1 & DownBlockx4 (Table \ref{tab:downblock}) & [B, 128, 32, 32] & 1,116,416 \\
     Layer3 & Layer2 & DownBlockx6 (Table \ref{tab:downblock}) & [B, 256, 16, 16] & 6,822,400 \\
     Layer4 & Layer3 & DownBlockx3 (Table \ref{tab:downblock}) & [B, 512, 8, 8] & 13,114,368 \\
     Out & Layer4 & Softmax & [B, 512, 8, 8] & 0\\
     \midrule
     \multicolumn{4}{r}{\textbf{Trainable Parameters:}} & \textbf{21,319,168} \\
     \bottomrule
\label{tab:RN34}
\end{tabular}
\endgroup
\end{table}

\begin{table}
\begingroup
\centering
\caption{ResNet18-Attn \protect\cite{he:15,zhang2019self}}
\small
\begin{tabular}{p{1.8cm} p{3.2cm} p{3.5cm} c r} 
    \toprule 
     Layer & Input & Operation & Output Shape & Parameters\\
     \midrule
     Conv1 & INPUT [B, 14, 128, 128] & Convolution2D & [B, 64, 64, 64] & 43,904 \\
     BatchNorm1 & Conv1 & Batch Normalisation & [B, 64, 64, 64] & 128 \\
     ReLu1 & BatchNorm1 & Rectified Linear Activation & [B, 64, 64, 64] & 0 \\
     Layer1 & ReLu1 & DownBlockx2 (Table \ref{tab:downblock}) & [B, 64, 64, 64] & 221,952 \\
     Layer2 & Layer1 & DownBlockx2 (Table \ref{tab:downblock}) & [B, 128, 32, 32] & 1,116,416 \\
     Layer3 & Layer2 & DownBlockx2 (Table \ref{tab:downblock}) & [B, 256, 16, 16] & 6,822,400 \\
     Attn1  & Layer2 & AttnBlock (Table \ref{tab:attnblock}) & [B, 256, 16, 16] & 82,241 \\
     Sum1   & [Attn1,Layer3] & Sum & [B, 256, 16, 16] & 0 \\
     Layer4 & Layer3 & DownBlockx2 (Table \ref{tab:downblock}) & [B, 512, 8, 8] & 13,114,368 \\
     Attn2  & Layer4 & AttnBlock (Table \ref{tab:attnblock}) & [B, 512, 8, 8] & 328,321 \\
     Sum2 & [Attn2, Layer4] & Sum & [B, 512, 8, 8] & 0 \\
     Out & Sum2 & Softmax & [B, 512, 8, 8] & 0 \\
     \midrule
     \multicolumn{4}{r}{\textbf{Trainable Parameters:}} & \textbf{11,621,570} \\
     \bottomrule
\label{tab:RN18Attn}
\end{tabular}
\endgroup
\end{table}

\begin{table}
\begingroup
\centering
\caption{Deconvolutional Header}
\small
\begin{tabular}{p{1.8cm} p{3.2cm} p{3.5cm} c r} 
    \toprule 
     Layer & Input & Operation & Output Shape & Parameters\\
     \midrule
     Layer1 & INPUT [B, 512, 8, 8] & UpBlockx1 (Table \ref{tab:upblock}) & [B, 256, 32, 32] & 1,705,728 \\
     Layer2 & Layer1 & UpBlockx1 (Table \ref{tab:upblock}) & [B, 128, 64, 64] & 426,880 \\
     Layer3 & Layer2 & UpBlockx1 (Table \ref{tab:upblock}) & [B, 53, 128, 128] & 106,944 \\
     Layer4 & Layer3 & UpBlockx1 (Table \ref{tab:upblock}) & [B, 32, 256, 256] & 26,848 \\
     Conv1 & Layer4 & Convolution2D$_{1x1}$  & [B, 6, 128, 128] & 231 \\
     \midrule
     \multicolumn{4}{r}{\textbf{Trainable Parameters:}} & \textbf{2,266,631} \\
     \bottomrule
\label{tab:DeconvHeader}
\end{tabular}
\endgroup
\end{table}

\section{Results}

\begin{figure}
\centering
\includegraphics[angle=90,width=1.0\textwidth]{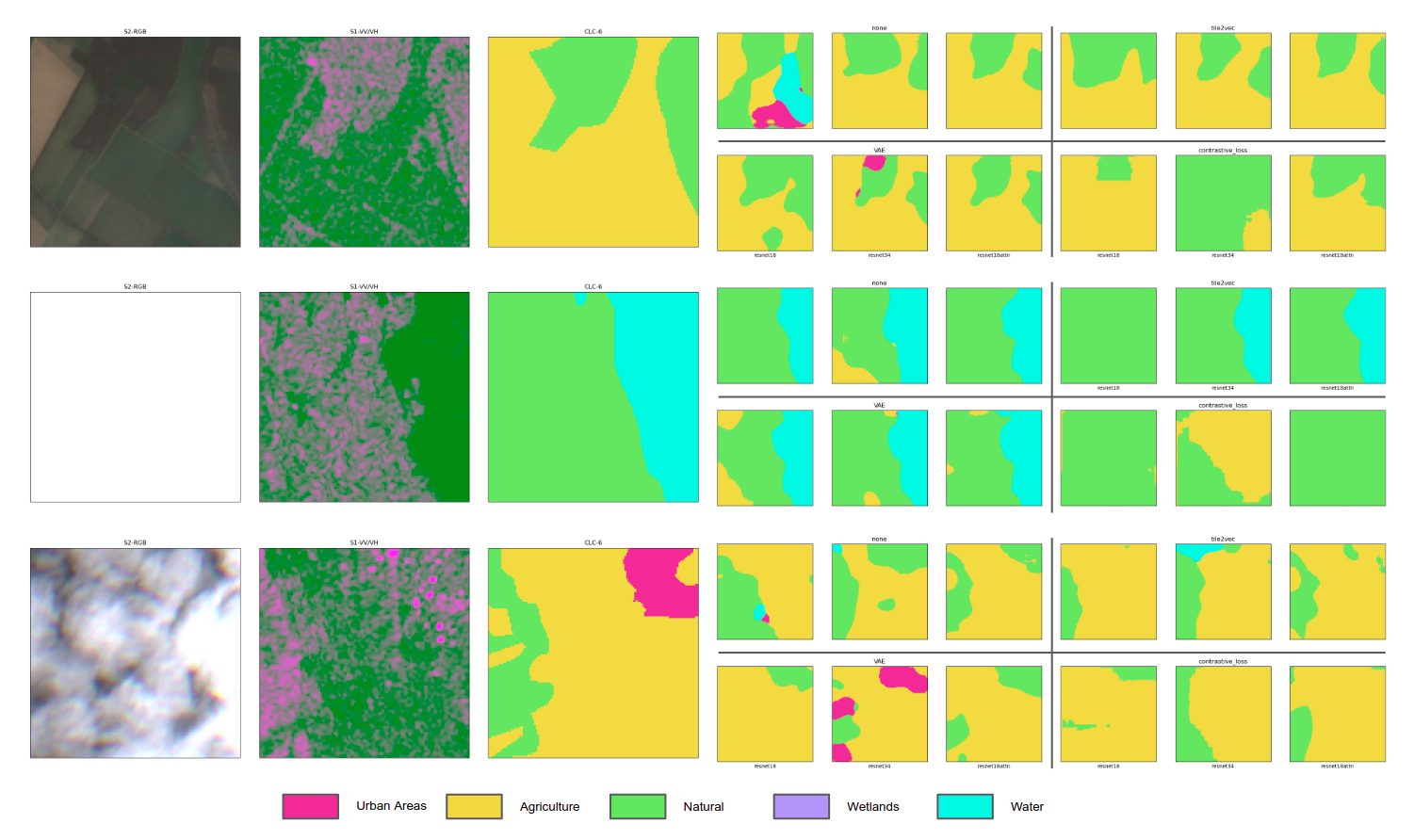} 
\caption{
Test Jaccard index (intersection-over-union) for 5 level-0 CORINE land cover classes  
}
\label{fig:results_samples}
\end{figure}

\begin{figure}
\centering
\includegraphics[width=1.15\textwidth]{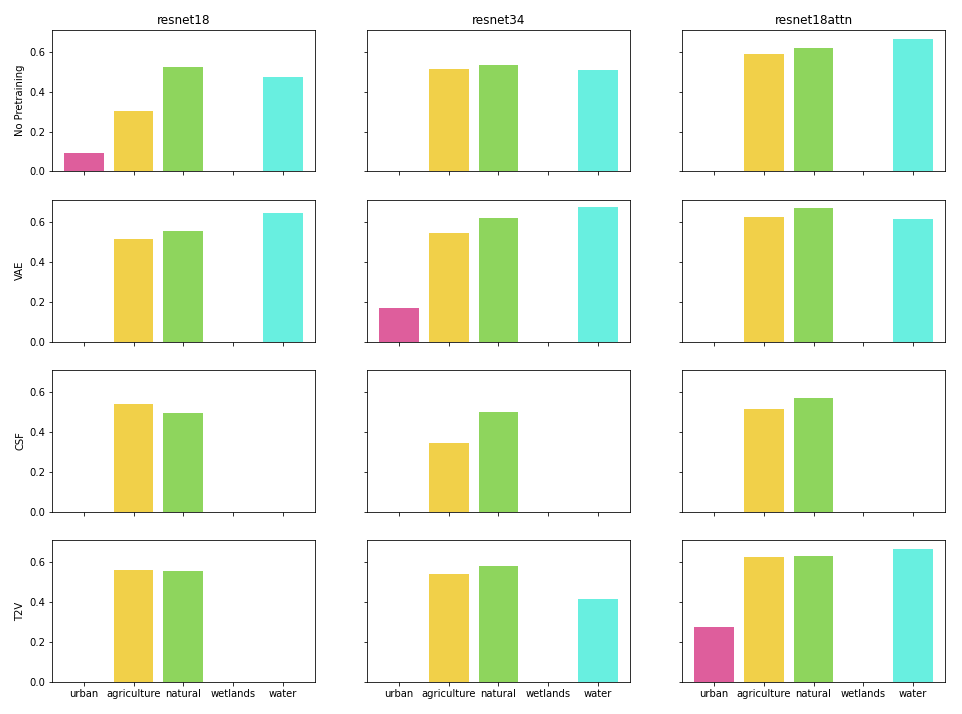} 
\caption{
Test Jaccard index (intersection-over-union) for 5 level-0 CORINE land cover classes  
}
\label{fig:CLC_IOU}
\end{figure}

Encoders and deconvolutional headers are trained for all combinations of encoders and pretaining algorithms.
Select results are shown in Figure \ref{fig:results_samples}.
The test performance of landcover classfication is shown in Table \ref{tab:LCClassification} and Figure \ref{fig:CLC_IOU}.
The Jaccard index (intersection-over-union, IoU) for each class is used as an evaluation metric. 
Table \ref{tab:LCClassification} shows IoU performance weighted by the size of each class. 
For all pretraining methods, the ResNet18 architecture with attention performed the best on the out-of-sample test set.
This suggests that parameter (and training data) size is not a key constraint in the development of useful embeddings, and that more can be done with novel architecture search.

Among the pretraining methods, variational autoencoding and tile2vec perform similarly well, while contrastive sensor fusion performs even worse than no pretraining with a ResNet34.
It is remarkable that the humble variational autoencoder was so competitive with the other pretraining methods.
Perhaps it is because the autoencoder must actively learn to encode levels of atmospheric interference in order to reconstruct it.

Figure \ref{fig:CLC_IOU} shows more detail into the performance of each pretraining and encoder combination for the 5 different land classes.
Many of the models struggled to classify urban areas, and none of them detected wetlands.
The models that were effective in classifying urban areas were three different combinations of encoder architecture and pretraining method, so a clear lesson is difficult to draw.
Perhaps the greater parameter space of ResNet34 allows for effectove autocoding of more sophisticated semantic classes.
Urban areas, with their distinct shapes, textures, and colors, may benefit more from attentions and spatially-explicit training, leading to the outperformance of tile2vec with ResNet18 with attention.

\begin{table}
\centering
\caption{Land Cover Classification, CLC level 0$^1$, test class-weighted mean IoU$^2$}
\begin{tabular}{p{2.2cm}|rrr} 
    \toprule
     Encoder & ResNet18 & ResNet34 & ResNet18-Attn \\
     Parameters$^3$ & 13.5mn  & 23.5mn  & 13.9mn  \\
     \midrule
     None & 38.5 & 48.2 & \textbf{55.7} \\
     VAE & 49.4 & \underline{54.8} & \textbf{59.4} \\
     T2V & \underline{50.2} & 51.3 & \underline{\textbf{59.7}} \\
     CSF & 46.3 & 37.6 & \textbf{48.8} \\
     \bottomrule
\end{tabular}
\label{tab:LCClassification}
{\newline\raggedright 1: Copernicus CORINE level 0 land cover classes, see Table \ref{tab:LCClasses}}
{\newline\raggedright 2: \textbf{bold} indicates top performing encoder by pretraining, \underline{underline} shows top performing pretraining by encoder.}
{\newline\raggedright 3: Includes encoder plus deconvolutional header}
\end{table}

\begin{table}
\centering
\caption{Copernicus CORINE Land Cover Classes}
\begin{tabularx}{7.2cm}{p{2.2cm} | p{5cm} } 
    \toprule
     Class Level 0 & Description \\
     \midrule
     1XX & Urban and built-up areas \\
     2XX & Agriculture and mixed land use \\
     3XX & Natural land uses \\
     4XX & Wetlands \\
     5XX & Permanent water \\
     \bottomrule
\end{tabularx}
\label{tab:LCClasses}
\end{table}

\section{Discussion}
While these results are novel, the poor performance suggests that more refinement is needed before this end-to-end self-supervised-plus-supervised method can be put into production systems. 
This machine learning task is difficult; atmospheric conditions obscure ground conditions, affect the brightness of the entire scene, and can cause artefacts like shadow.
Figure \ref{fig:1-examples} shows how much atmospheric interference is present in this `raw' data, with a very small portion of the scenes being cloud-free, and many scenes being entirely saturated by cloud cover.
Atmosphere conditions also affect SAR imagery, albeit to a lesser extent.
Target distributions are dynamic, changing (marginally) through time and also cyclically with the seasons and natural growing cycles.
Training labels are up-scaled from a nominally coarser resolution and so are `noisy', further compounding the challenge of the learning task.

It is also quite likely that the computer vision models employed in this paper are not fit for remote sensing tasks.
Remote sensing imagery has several properties which make it distinctly different from the general imagery (as might be captured by a phone or dashboard camera) that these models were developed for.
Some intuitions are accessible:
\begin{itemize}
    \item Remote sensing imagery has no horizon and fixed focal distance. Unlike general-purpose encoders, those for remote sensing do not need to codify perspective, angle, and scale.
    \item Objects of interest are often fewer pixels in size than in conventional computer vision problems. This means information does not need to traverse large distances in the pixel space to inform a classification or regression. Conventional computer vision architectures trade off pixel-space for feature space and use pooling compress representations. This might not be necessary with remote sensing imagery.
    \item Remote sensing imagery is more often confounded by atmospheric conditions. This means surface representations are more variant and models need to be designed to accomodate this dynamicism and uncertainty.
    \item Often more spectra are available than the red-green-blue optical spectra of conventional imagery. This means there is combinationally more complexity accessible in the lower latent space encodings which these models have not been designed to capture.
\end{itemize}

Some ideas for future research might include exploring the trade-off between contextual information filtered with attention and local information. 
Grouping convolutions might help draw out relations between a high number of input bands.
Multi-look training, especially with SAR data, might help normalise data within a single scene.
Newer, shallower architectures might be tried to more fully encode the shallow latent space.
New methods for modelling uncertainty might help isolate the effect of atmospheric conditions on model predictions.
DeepSentinel provides a platform for this future research.

\section{Conclusions}

This paper presents DeepSentinel, a new codebase and dataset for Sentinel-1, Sentinel-2 sensor fusion research and applications.
A preliminary experiment with DeepSentinel is developed, testing a number of encoder architectures and pretraining methods for accuracy on a landcover classification task.
Of the encoder architrectures and pretraining methods trialed, tile2vec pretraining with a shallow ResNet variant with self-attention is found to produce the best classifier in intersection-over-union tests against held-out data.
The mixed results help give intuition for future research of this type, particularly the novelties of earth observation data.
DeepSentinel is an open-source tool and dataset which can facilitate these future efforts.

\begin{ack}
I am immensely grateful for the ongoing support of Descartes Labs who have provided platform access for this project as part of their Impact Science Program. This work is also being prepared with the generous support of Microsoft AI for Earth and Google Cloud Platform research credits. Without the support of these organisations this work would not be possible; they have my immense gratitude.
\end{ack}

\medskip

\small

\bibliography{main}

\begin{thebibliography}{10}

\bibitem{descarteslabs17}
C.~M. {Beneke}, S.~{Skillman}, M.~S. {Warren}, T.~{Kelton}, S.~P. {Brumby},
  R.~{Chartrand}, and M.~{Mathis}.
\newblock {A Platform for Scalable Satellite and Geospatial Data Analysis}.
\newblock In {\em AGU Fall Meeting Abstracts}, volume 2017, pages IN32C--04,
  Dec. 2017.

\bibitem{bermudez2018}
J.~Bermudez, P.~Happ, D.~Oliveira, and R.~Feitosa.
\newblock Sar to optical image synthesis for cloud removal with generative
  adversarial networks.
\newblock {\em ISPRS Annals of Photogrammetry, Remote Sensing \& Spatial
  Information Sciences}, 4(1), 2018.

\bibitem{bermudez2019}
J.~D. Bermudez, P.~N. Happ, R.~Q. Feitosa, and D.~A. Oliveira.
\newblock Synthesis of multispectral optical images from sar/optical
  multitemporal data using conditional generative adversarial networks.
\newblock {\em IEEE Geoscience and Remote Sensing Letters}, 16(8):1220--1224,
  2019.

\bibitem{spglobal20}
B.~Burks.
\newblock {Space, The Next Frontier: Spatial Finance And Environmental
  Sustainability}.
\newblock
  \texttt{https://www.spglobal.com/ratings/en/research/articles/200122-space-the-next-frontier-spatial-finance-and-environmental-sustainability-11317146},
  2020.

\bibitem{caldecott2018}
B.~Caldecott, L.~Kruitwagen, M.~McCarten, X.~Zhou, D.~Lunsford, O.~Marchand,
  P.~Hadjikyriakou, V.~Bickel, T.~Sachs, and N.~Bohn.
\newblock Climate risk analysis from space: remote sensing, machine learning,
  and the future of measuring climate-related risk, 2018.

\bibitem{eckardt2013}
R.~Eckardt, C.~Berger, C.~Thiel, and C.~Schmullius.
\newblock Removal of optically thick clouds from multi-spectral satellite
  images using multi-frequency sar data.
\newblock {\em Remote Sensing}, 5(6):2973--3006, 2013.

\bibitem{ferrant2017}
S.~Ferrant, A.~Selles, M.~Le~Page, P.-A. Herrault, C.~Pelletier, A.~Al-Bitar,
  S.~Mermoz, S.~Gascoin, A.~Bouvet, M.~Saqalli, et~al.
\newblock Detection of irrigated crops from sentinel-1 and sentinel-2 data to
  estimate seasonal groundwater use in south india.
\newblock {\em Remote Sensing}, 9(11):1119, 2017.

\bibitem{fuentes2019}
M.~Fuentes~Reyes, S.~Auer, N.~Merkle, C.~Henry, and M.~Schmitt.
\newblock Sar-to-optical image translation based on conditional generative
  adversarial networks—optimization, opportunities and limits.
\newblock {\em Remote Sensing}, 11(17):2067, 2019.

\bibitem{gee17}
N.~Gorelick, M.~Hancher, M.~Dixon, S.~Ilyushchenko, D.~Thau, and R.~Moore.
\newblock Google earth engine: Planetary-scale geospatial analysis for
  everyone.
\newblock {\em Remote Sensing of Environment}, 2017.

\bibitem{cti18}
M.~Gray, L.~Watson, S.~Ljungwaldh, and E.~Morris.
\newblock Nowhere to hide: Using satellite imagery to estimate the utilisation
  of fossil fuel power plants, 2018.

\bibitem{grohnfeldt2018}
C.~Grohnfeldt, M.~Schmitt, and X.~Zhu.
\newblock A conditional generative adversarial network to fuse sar and
  multispectral optical data for cloud removal from sentinel-2 images.
\newblock In {\em IGARSS 2018-2018 IEEE International Geoscience and Remote
  Sensing Symposium}, pages 1726--1729. IEEE, 2018.

\bibitem{ha2018world}
D.~Ha and J.~Schmidhuber.
\newblock World models.
\newblock {\em arXiv preprint arXiv:1803.10122}, 2018.

\bibitem{he:15}
K.~He, X.~Zhang, S.~Ren, and J.~Sun.
\newblock Deep residual learning for image recognition.
\newblock {\em CoRR}, abs/1512.03385, 2015.

\bibitem{he2018}
W.~He and N.~Yokoya.
\newblock Multi-temporal sentinel-1 and-2 data fusion for optical image
  simulation.
\newblock {\em ISPRS International Journal of Geo-Information}, 7(10):389,
  2018.

\bibitem{jean2019}
N.~Jean, S.~Wang, A.~Samar, G.~Azzari, D.~Lobell, and S.~Ermon.
\newblock Tile2vec: Unsupervised representation learning for spatially
  distributed data.
\newblock In {\em Proceedings of the AAAI Conference on Artificial
  Intelligence}, volume~33, pages 3967--3974, 2019.

\bibitem{kingma2013auto}
D.~P. Kingma and M.~Welling.
\newblock Auto-encoding variational bayes.
\newblock {\em arXiv preprint arXiv:1312.6114}, 2013.

\bibitem{kruitwagen19}
L.~Kruitwagen, K.~Stoy, J.~Friedrich, S.~Skillman, and C.~Hepburn.
\newblock A global census of solar facilities using deep learning and remote
  sensing.
\newblock In {\em NeurIPS Climate Change AI Workshop 2019}. NeurIPS Climate
  Change AI, 2019.

\bibitem{meraner2020}
A.~Meraner, P.~Ebel, X.~X. Zhu, and M.~Schmitt.
\newblock Cloud removal in sentinel-2 imagery using a deep residual neural
  network and sar-optical data fusion.
\newblock {\em ISPRS Journal of Photogrammetry and Remote Sensing},
  166:333--346, 2020.

\bibitem{mikolov2013efficient}
T.~Mikolov, K.~Chen, G.~Corrado, and J.~Dean.
\newblock Efficient estimation of word representations in vector space.
\newblock {\em arXiv preprint arXiv:1301.3781}, 2013.

\bibitem{ndikumana2018}
E.~Ndikumana, D.~Ho~Tong~Minh, N.~Baghdadi, D.~Courault, and L.~Hossard.
\newblock Deep recurrent neural network for agricultural classification using
  multitemporal sar sentinel-1 for camargue, france.
\newblock {\em Remote Sensing}, 10(8):1217, 2018.

\bibitem{nelson2014}
A.~Nelson, T.~Setiyono, A.~B. Rala, E.~D. Quicho, J.~V. Raviz, P.~J. Abonete,
  A.~A. Maunahan, C.~A. Garcia, H.~Z.~M. Bhatti, L.~S. Villano, et~al.
\newblock Towards an operational sar-based rice monitoring system in asia:
  Examples from 13 demonstration sites across asia in the riice project.
\newblock {\em Remote Sensing}, 6(11):10773--10812, 2014.

\bibitem{oord2018representation}
A.~v.~d. Oord, Y.~Li, and O.~Vinyals.
\newblock Representation learning with contrastive predictive coding.
\newblock {\em arXiv preprint arXiv:1807.03748}, 2018.

\bibitem{osm}
{OpenStreetMap contributors}.
\newblock {Planet dump retrieved from https://planet.osm.org }.
\newblock \texttt{ https://www.openstreetmap.org }, 2020.

\bibitem{ruetschi2018}
M.~R{\"u}etschi, M.~E. Schaepman, and D.~Small.
\newblock Using multitemporal sentinel-1 c-band backscatter to monitor
  phenology and classify deciduous and coniferous forests in northern
  switzerland.
\newblock {\em Remote Sensing}, 10(1):55, 2018.

\bibitem{corine:20}
C.~L.~M. Service.
\newblock Corine land cover, 2020.

\bibitem{usda:20}
N.~A.~S. Service.
\newblock Cropland data layer, 2020.

\bibitem{setiyono2017}
T.~Setiyono, F.~Holecz, N.~Khan, M.~Barbieri, E.~Quicho, F.~Collivignarelli,
  A.~Maunahan, L.~Gatti, and G.~Romuga.
\newblock Synthetic aperture radar (sar)-based paddy rice monitoring system:
  Development and application in key rice producing areas in tropical asia.
\newblock In {\em IOP Conference Series: Earth and Environmental Science},
  volume~54, page 012015. IOP Publishing, 2017.

\bibitem{swope2020representation}
A.~M. Swope, X.~H. Rudelis, and K.~T. Story.
\newblock Representation learning for remote sensing: An unsupervised sensor
  fusion approach, 2020.

\bibitem{tian2019contrastive}
Y.~Tian, D.~Krishnan, and P.~Isola.
\newblock Contrastive multiview coding.
\newblock {\em arXiv preprint arXiv:1906.05849}, 2019.

\bibitem{tobler1970computer}
W.~R. Tobler.
\newblock A computer movie simulating urban growth in the detroit region.
\newblock {\em Economic geography}, 46(sup1):234--240, 1970.

\bibitem{torbick2018}
N.~Torbick, X.~Huang, B.~Ziniti, D.~Johnson, J.~Masek, and M.~Reba.
\newblock Fusion of moderate resolution earth observations for operational crop
  type mapping.
\newblock {\em Remote Sensing}, 10(7):1058, 2018.

\bibitem{van2018}
K.~Van~Tricht, A.~Gobin, S.~Gilliams, and I.~Piccard.
\newblock Synergistic use of radar sentinel-1 and optical sentinel-2 imagery
  for crop mapping: a case study for belgium.
\newblock {\em Remote Sensing}, 10(10):1642, 2018.

\bibitem{varon2020}
D.~J. Varon, D.~J. Jacob, D.~Jervis, and J.~McKeever.
\newblock Quantifying time-averaged methane emissions from individual coal mine
  vents with ghgsat-d satellite observations.
\newblock {\em Environmental Science \& Technology}, 54(16):10246--10253, 2020.

\bibitem{vaswani2017attention}
A.~Vaswani, N.~Shazeer, N.~Parmar, J.~Uszkoreit, L.~Jones, A.~N. Gomez,
  L.~Kaiser, and I.~Polosukhin.
\newblock Attention is all you need.
\newblock {\em arXiv preprint arXiv:1706.03762}, 2017.

\bibitem{zhang2019self}
H.~Zhang, I.~Goodfellow, D.~Metaxas, and A.~Odena.
\newblock Self-attention generative adversarial networks.
\newblock In {\em International conference on machine learning}, pages
  7354--7363. PMLR, 2019.

\bibitem{zhang2019}
Q.~Zhang, Q.~Kong, C.~Zhang, S.~You, H.~Wei, R.~Sun, and L.~Li.
\newblock A new road extraction method using sentinel-1 sar images based on the
  deep fully convolutional neural network.
\newblock {\em European Journal of Remote Sensing}, 52(1):572--582, 2019.

\end{thebibliography}

\end{document}